\documentclass[preprint,12pt]{elsarticle}

\usepackage{amssymb}
\usepackage{amsmath}
\usepackage{xurl}
\usepackage{hyperref}
\usepackage{booktabs}
\usepackage{xcolor}%
\usepackage{soul}

\usepackage{multirow}

\journal{Nuclear Physics B}

\begin{document}

\begin{frontmatter}



\title{Polarity Detection of Sustainable Development Goals in News Text}

\author{Andrea Cadeddu\fnref{label2}}
\ead{andrea.cadeddu@linkalab.it}
\author{Alessandro Chessa\fnref{label2}}
\ead{alessandro.chessa@linkalab.it}
\author{Vincenzo De Leo\fnref{label2,label3}}
\ead{vincenzo.deleo@linkalab.it}
\author{Gianni Fenu\fnref{label3}}
\ead{fenu@unica.it}
\author{Francesco Osborne\fnref{label4}}
\ead{francesco.osborne@open.ac.uk} 
\author{Diego Reforgiato Recupero\corref{cor1}\fnref{label3}}
\ead{diego.reforgiato@unica.it}
\author{Angelo Salatino\fnref{label4}}
\ead{angelo.salatino@open.ac.uk} 
\author{Luca Secchi\fnref{label2}}
\ead{luca.secchi@linkalab.it}


\cortext[cor1]{Corresponding Author}
\affiliation[label2]{organization={Linkalab s.r.l.},
             addressline={Viale Elmas, 142},
             city={Cagliari},
             postcode={09122},
             country={Italy}}
\affiliation[label3]{organization={Department of Mathematics and Computer Science, University of Cagliari},
             addressline={via Ospedale 72},
             city={Cagliari},
             postcode={09124},
             country={Italy}}
\affiliation[label4]{organization={Knowledge Media Institute, The Open University},
             addressline={Walton Hall, Kents Hill},
             city={Milton Keynes},
             postcode={MK76AA},
             country={United Kingdom}}

\begin{abstract}

The United Nations’ Sustainable Development Goals (SDGs) provide a globally recognised framework for addressing major societal, environmental, and economic challenges. While recent advances in natural language processing (NLP) and large language models (LLMs) have enabled the automatic identification of SDG-related content, they do not capture whether the described events represent progress toward or regression from a specific goal. To address this gap, we introduce the novel task of SDG polarity detection and present SDG-POD, a benchmark dataset combining manually annotated and synthetically generated examples. We evaluate six state-of-the-art open-source LLMs under both zero-shot and fine-tuning settings and investigate the impact of synthetic data augmentation on model performance. Our results show that SDG polarity detection remains challenging for current LLMs; however, fine-tuned models, particularly QWQ-32B, achieve the best overall performance, with especially strong results on SDG-9, SDG-12, and SDG-15. Furthermore, we demonstrate that synthetic training data consistently improves model robustness and classification performance. This work introduces a new benchmark for SDG polarity detection and provides practical insights into developing LLM-based systems for sustainability monitoring.

\end{abstract}




\end{frontmatter}



\section{Introduction}

The 17 Sustainable Development Goals (SDGs) form the cornerstone of the 2030 Agenda for Sustainable Development, unanimously adopted by all United Nations Member States in 2015. They address critical global challenges, including poverty alleviation, hunger eradication, quality education, gender equality, climate action, responsible consumption, and ecosystem preservation.  These goals currently represent the most comprehensive and widely endorsed framework designed to measure and advance global progress toward sustainability across social, economic, and environmental dimensions.


Tracking and assessing progress towards achieving the United Nations' SDGs currently presents a considerable challenge due to the vastness and complexity of relevant data~\citep{HIRAI2022109605, RePEc:gam:jsusta:v:15:y:2023:i:4:p:3203-:d:1063481, Mishra2023, LaFleur2023}\footnote{\url{https://unece.org/sites/default/files/2021-04/2012761_E_web.pdf}}. Traditional manual analysis techniques can no longer cope with the rapid, large-scale data generation characteristic of today's interconnected global environment~\citep{Nilashi2023Critical}.

In response, machine learning models have emerged as indispensable tools capable of processing and analysing extensive textual data from diverse sources, including institutional reports, news articles, social media platforms, and official documents~\citep{10.1007/978-3-031-21743-2_21, Matsui2022, Guisiano2021}. These models facilitate the automatic detection and categorization of content associated with individual SDGs, thereby supporting real-time monitoring and evaluation of the effectiveness of SDG-related initiatives.

Most research in this area primarily focuses on determining whether a given text segment is related to a specific SDG. This task is typically formalized as a binary classification problem, wherein the objective is to assess whether a text pertains to a particular SDG. However, merely associating a text segment with an SDG often lacks the granularity needed to fully understand the implications and broader context of the described activities or events. For example, a text discussing a new policy designed to reduce hunger in developing regions and another describing an emerging famine crisis would both be associated with SDG 2 ("Zero Hunger"), despite reflecting fundamentally different impacts on sustainability goals: one potentially positive and the other clearly negative.

Consequently, binary classification alone proves insufficient for many practical purposes, as it does not capture the direction or nature of the impact described. In numerous real-world scenarios, it is essential to further analyse whether the activities or events described in the texts have positive, neutral, or negative implications for progress toward achieving the respective SDGs. This more detailed analysis can significantly enhance researchers' and policymakers' abilities to evaluate and respond appropriately to developments affecting global sustainability objectives~\citep{Nilashi2023Critical}. 
This task aligns with a well-established problem in Natural Language Processing (NLP) known as \textit{polarity detection}~\citep{bianchi-etal-2021-sweat,Nemeth2023scoping}. In this context, polarity detection refers to identifying the effect (positive, neutral, or negative) of the actions or activities described in a text with respect to a specific SDG.  
It is crucial to note that this task is distinct from traditional sentiment analysis~\citep{10.3115/1118693.1118704}, which aims instead to determine the overall positive or negative tone of a text, regardless of its relevance to any particular SDG. While sentiment analysis focuses on subjective opinions or emotional tone, polarity detection in the context of SDGs is concerned with evaluating the actual impact of described actions in relation to sustainability targets (i.e., whether a given text expresses evidence of progress toward a specific SDG or conveys an intention to pursue such progress). As we will discuss in more detail in Section \ref{sec:Task}, it is therefore possible for a text to exhibit a positive sentiment while conveying a negative polarity, or vice versa.

Therefore, in this paper, we use the term SDG polarity to denote the direction of the impact that a text describes with respect to a specific SDG, namely whether it indicates progress towards the goal, regression from it, or neither. This notion should not be confused with political or ideological polarization, nor with conventional sentiment analysis. Instead, it represents a task-specific semantic assessment of the relationship between a textual statement and an SDG.

Despite its evident importance, polarity detection in the context of SDGs remains largely underexplored, primarily due to the absence of comprehensive, annotated datasets necessary to effectively train and evaluate relevant NLP models. Addressing this data gap constitutes a critical step toward enabling more detailed, actionable analyses of texts concerning sustainable development.
Concurrently, recent progress in NLP has largely been propelled by the introduction and advancement of large language models (LLMs), which have established new benchmarks across diverse tasks~\citep{10.5555/3295222.3295349}. These models demonstrate exceptional capabilities in capturing intricate linguistic structures and semantic nuances, achieving superior results in tasks including text classification, sentiment analysis, polarity detection, and general language comprehension~\citep{devlin-etal-2019-bert, NEURIPS2020_1457c0d6}. 

In this paper, we address the task of SDG polarity detection by introducing a novel benchmark and conducting a comprehensive evaluation of LLMs. Specifically, we present SDG-POD (SDG POlarity Detection), a new benchmark dataset developed to support the training and evaluation of LLMs for this task. The dataset contains 6,400 texts, each annotated with a polarity label. We evaluate six state-of-the-art LLMs on the SDG-POD dataset under both zero-shot (ZSL) and fine-tuned conditions. Furthermore, we investigate the effect of augmenting the fine-tuning data with synthetic examples generated by multiple LLMs. To enhance the quality of these examples, we apply a majority voting strategy to select the most reliable synthetic instances, thereby improving the robustness of the training set. \color{black}In our experiments, we exclusively employed open‐source models rather than proprietary alternatives, since the study’s primary aim was to evaluate the performance of this class of LLMs on the specified task and within the designated domain. Specifically, we conducted all experiments using a diverse set of models, ranging from BERT variants with only a few hundred million parameters to the QWQ-2.5 model with 32 billion parameters.

Our results suggest that the task remains challenging for the current generation of LLMs.  This difficulty stems from the subtle and abstract nature of the text, which human experts can interpret effectively but which remains much more challenging for LLMs. Consequently, the SDG-POD represents a significant challenge for future research and models. We hope that the community will continue to adapt and make progress on this task. 
\color{black}
Nevertheless, some fine-tuned models, particularly QWQ-32B, achieve fair performance, especially on specific Sustainable Development Goals (SDGs) such as SDG-9 (Industry, Innovation and Infrastructure), SDG-12 (Responsible Consumption and Production), and SDG-15 (Life on Land). 
Moreover, our analysis underscores the effectiveness of data enrichment techniques in mitigating the difficulties inherent to this domain, which is characterized by limited resources for training.


In summary, the main contributions of this paper are as follows:
\begin{itemize}
\item SDG-POD, a novel benchmark for polarity detection in the context of the SDGs, which integrates both manually annotated and synthetically generated data;
\item A comparative evaluation of several LLMs on SDG-POD, considering both ZSL and fine-tuning settings;
\item An investigation into the use of synthetic data, generated through an agentic architecture employing five LLMs, to enhance the performance of polarity detection models;
\item The complete codebase of the experiments, made publicly available to facilitate reproducibility and enable the community to reuse the trained models\footnote{Codebase of the experiments: \url{https://github.com/vincenzodeleo/sdg_polarity_detection/tree/main}}.
\end{itemize}

The remainder of this paper is organized as follows. Section \ref{sec:Relatedwork} reviews the related literature. In Section \ref{sec:Task}, we describe the specific task addressed in this study. Section \ref{sec:Benchmark} introduces the proposed benchmark. Section \ref{sec:Methodology} outlines the experimental methodology. The results are presented and analyzed in Section \ref{sec:results}. Finally, Section \ref{sec:conclusions} concludes the paper and offers recommendations for researchers seeking to develop solutions that effectively balance performance and resource efficiency.


\section{Related Work}\label{sec:Relatedwork}

Sentiment analysis~\citep{10.3115/1118693.1118704}, which involves understanding the emotional tone of a text to determine whether it expresses a positive, negative, or neutral sentiment, is one of the most dynamic research lines in the field of Natural Language Processing (NLP). 
Initial approaches to sentiment analysis often relied on techniques based on predefined lexicons of positive/negative terms or on traditional machine learning algorithms (e.g., logistic regression, Naïve Bayes, Support Vector Machines) applied on manually extracted features from text~\citep{Tan2023Survey}. Such conventional methods have shown limitations in capturing the linguistic nuances and complex context of natural texts. 

A key advance has been the introduction of Transformer-based models~\citep{10.5555/3295222.3295349}, in particular pre-trained Language Models such as BERT~\citep{devlin-etal-2019-bert} and its derivatives~\citep{conf/emnlp/XiaWD20}, which have rapidly advanced the state of the art in sentiment analysis~\citep{Alaparthi2021, 10.1007/978-3-030-86472-9_13}. Transformer models, thanks to the attention mechanism, can effectively capture long-range relationships in text and their fine-tuning on sentiment datasets has led to state-of-the-art results on many sentiment analysis benchmarks, often with a significant margin over previous methods~\citep{Wu2024}. For example, Kokab et al.~\citep{Kokab2022Transformer} proposed a generalized Transformer-based model for sentiment analysis on social media data, capable of handling noisy texts, out-of-vocabulary words, and context loss. 

A very recent emerging trend is the application of generative LLMs, such as GPT~\citep{openai2024gpt4technicalreport} or LLaMA models~\citep{grattafiori2024llama3herdmodels}, to the task of sentiment analysis. These models, trained on huge amounts of generic text data, demonstrate a surprising ability to generalize to specific tasks via simple prompting (i.e., instructions in natural language) without the need for further supervised training. Krugmann and Hartmann~\citep{Krugmann2024LLM} performed a pioneering study in which they directly compared the ability of generative LLM models to classify sentiment with traditional specialized transfer learning models. The results show that the latest LLMs, despite operating in zero-shot mode (i.e., without having been specifically trained on the target dataset), can match and sometimes exceed the performance of conventional fine-tuned models on sentiment data, in terms of classification accuracy. This represents a paradigm shift: general-purpose generative models can perform sentiment analysis tasks with comparable effectiveness to that of purpose-built models. 

As an extension of sentiment analysis, another task has recently begun to emerge with increasing interest, called stance detection~\citep{10.1145/3404835.3462815}, which, instead of analyzing the simple emotional tone of a text (analyzed by sentiment analysis), deals with determining the explicit or implicit position expressed towards a specific target. It differs from sentiment analysis because it does not necessarily coincide with the emotional tone of a text: for example, a tweet with a positive tone can still oppose a certain proposal, and vice versa~\citep{AlDayel2021}. 
Initially, the best results in stance detection were obtained with traditional supervised classifiers~\citep{Alturayeif2023} (e.g., SVM or logistic regressors) based on manual features such as discriminant terms, n-grams, and lexical indicators. Early approaches with neural networks on short and noisy data (tweets) proved to be less effective, due to the scarcity of annotated data and the informal language of social media~\citep{ALDAYEL2021102597}. In recent years, however, the adoption of pre-trained deep learning models has led to significant progress. In particular, Transformer models (e.g., BERT, RoBERTa) fine-tuned on specific collections have rapidly outperformed manual feature models, becoming the standard for stance detection on texts~\citep{Alturayeif2023}.

Although these three research areas are closely related, they address different objectives. Sentiment analysis aims to identify the emotional tone expressed in a text, stance detection determines whether the author supports or opposes a specific target, while polarization analysis typically studies ideological or opinion-based divisions within a collection of texts. In contrast, the task investigated in this paper—SDG polarity detection—does not seek to infer emotions, opinions, or ideological positions. Instead, it evaluates whether the events or actions described in a text indicate progress towards, regression from, or no meaningful effect on a specific Sustainable Development Goal.

In addition to sentiment analysis and stance detection, there is a third type of analysis capable of providing information on a dataset from the point of view of the degree to which the contents of a text (or a set of texts) are divided into opposing or extreme positions, named polarity detection. In other words, it involves evaluating how strongly a text expresses polarized positions, that is, clearly positive vs. negative, favorable vs. contrary, or adhering to opposing sides on a topic. This concept is applicable in any context (political, media, academic, social, etc.) and indicates the presence of a clear dichotomy in the opinions or feelings expressed in written language. A highly polarized textual discussion tends to cluster around two (or more) distinct ideological or emotional poles, with little presence of moderate or neutral tones in the middle. Textual polarization analysis, therefore, consists of identifying and measuring these divisions within the language, highlighting how much the texts are unbalanced towards opposing extremes rather than distributed along intermediate positions~\citep{bianchi-etal-2021-sweat}. In recent years, research on the analysis of polarization in texts has seen an increasing use of deep learning techniques and Transformers-based models to detect and quantify ideological divisions in linguistic content. A recent review study has highlighted how over a third of works in the last two years adopt machine learning approaches for this purpose~\citep{Nemeth2023scoping}.

Pre-trained Transformer models such as BERT and RoBERTa have become key tools to classify the ideological orientation or pole of a text document~\citep{Chriqui_2022}. One of the most well-known research strands of this type of analysis focuses on the automatic identification of the political position expressed in social media or news texts, treating the problem as a variant of text classification~\citep{https://doi.org/10.1002/aaai.12104}. For example, a BERT-based model has been proposed to detect political ideology in tweets about COVID-19, as an indicator of polarization in discussions about vaccines and masks~\citep{Kabir2022ideology}. In this study, the use of BERT allows to capture the linguistic context of tweets, while the integration of emotional signals improved the accuracy in distinguishing conservative vs. progressive users. In parallel, other works combine sentiment and emotion analysis with deep learning models to identify forms of affective polarization (i.e. hostility towards the opposite group); these approaches recognize that expressions of strong positive/negative sentiment towards an issue can signal polarized positions~\citep{10.1145/3041021.3054223}.

In the last decade, since the definition of SDG in 2015, several scientific works have been published related to the SDG domain from various perspectives. For example, Schmidt-Traub et al.~\citep{Schmidt-Traub2017} introduced the SDG Index as an analytical tool to assess countries' baselines for the SDGs, which can be applied by researchers in the interdisciplinary analyses needed for implementation. Similarly, Vanderfeesten et al.~\citep{Vanderfeesten2022} developed the well-known ``SDG Classifier'', a tool that allowed to automatically map the scientific literature to the SDGs, based on BERT\footnote{\url{https://zenodo.org/records/6487606}}. At the same time, Pradhan et al.~\citep{https://doi.org/10.1002/2017EF000632} studied and identified the interactions between different SDGs, to discover synergies and trade-offs between several SDG pairs. Rosemberg et al.~\citep{ROSENBERG2023101287} instead developed the sentiment analysis of Twitter data on climate change, which is transversal to several SDGs. Gennari and D'Orazio~\citep{doi:10.3233/SJI-200688} finally created an approach based on statistical methods to evaluate the progress towards the sustainable development goals.

Another relevant initiative is the STRINGS (Steering Research and Innovation for Global Goals) project\footnote{\url{https://strings.org.uk/}}, which proposes an integrative framework for mapping research areas to the SDGs while explicitly recognising that the relationship between research and SDGs is often open to multiple interpretations. Rather than assigning a single definitive mapping, the STRINGS framework encourages stakeholders to interpret the relevance of research according to different perspectives and provides visual tools for exploring these relationships~\cite{repec:osf:socarx:yfqbd}. Our work is complementary to this line of research. Whereas STRINGS focuses on identifying and representing the relationship between research and SDGs, SDG polarity detection aims to characterize the direction of that relationship by determining whether the text describes progress towards, regression from, or no meaningful impact on a specific SDG.

Despite significant advances in the fields of sentiment analysis, stance detection and polarization analysis, to date no study has specifically addressed the concept of ``SDG polarization'' as a polarity detection task. In other words, there is still no analysis that verifies whether a text refers to events or intentions indicative of an advancement or not with respect to a specific sustainable development goal. The few similar works found in the literature concern, for example, the use of advanced NLP methods to identify how much companies contribute positively to the SDGs by analyzing the texts of their sustainability reports (CSR)~\citep{Chen2022}. In this study, the sustainability reports of 1000 companies (2010–2019) were used to train classification models (logistics, SVM, and neural network) in order to predict the alignment of each company to the 17 SDGs. The approach combines thematic dictionaries on SDGs with word embeddings (Word2Vec, Doc2Vec) to represent the text, improving the classification performance. The best model (SVM with Doc2Vec embedding) achieves over 80\% accuracy in distinguishing whether a company is aligned with the SDGs (therefore indicating actions progressing towards the goals). Other works, such as that of Funk et al.~\citep{Funk2024}, apply Aspect-Based Sentiment Analysis (ABSA) to Voluntary National Reviews (VNR), the reports with which 166 countries periodically describe their progress on the SDGs, to measure Sentiment in national reports on SDG progress. The method extracts for each country a sentiment score for each of the 17 SDGs, indicating how positively (or negatively) the country speaks of progress on each goal; these textual scores (SDG sentiment score) are then compared with official UN indicators. The reported results showed that for most of the SDGs the correlation is not significant, indicating that the contents of the VNRs do not always reflect the real trend.

For this reason, we developed an innovative analysis based on the use of LLMs, described and explored in the following sections, aimed at developing automatic analysis systems capable of supporting government bodies and companies in measuring the progress of projects and monitoring the activities of the various actors involved in achieving the objectives of the UN Agenda 2030.

\section{Task Definition}\label{sec:Task}

In this paper, we formalise the novel task of SDG polarity detection as a single-label, multi-class classification problem. Here, the term polarity refers exclusively to the direction of the impact that the content of a text has with respect to a given SDG. Specifically, the objective is to determine whether a text describes evidence of progress towards the SDG (positive), regression from the SDG (negative), or neither (neutral). This definition is distinct from both sentiment analysis and political or ideological polarization.

The three possible class values of this task are:

\begin{itemize}
    \item \textbf{Positive polarity}: Represents the case in which the text contains indications related to a progress state or the intention to bring about a progress state with respect to the specific SDG provided alongside the text.  For example, a document describing a new initiative aimed at promoting gender equality would be classified as having positive polarity with respect to SDG~5.
    \item \textbf{Neutral polarity}: Represents the case in which the text does not provide evidence of progress towards or regression from the specified SDG. This category includes descriptive, analytical, or monitoring texts that discuss an SDG-related issue without implying that the reported events or actions contribute positively or negatively to the achievement of the goal.
    \item \textbf{Negative polarity}: Represents the case in which the text contains indications related to a state of regression or the intention to bring about a state of regression with respect to the specific SDG provided alongside the text. For example, a text reporting an increase in global famine would exhibit negative polarity with respect to SDG~2.
\end{itemize}

\color{black}
While classifying a text according to the relevant SGD is a largely solved task, the analysis presented in this paper demonstrates that detecting the polarity of SDG remains a surprisingly difficult challenge, even for the current generation of LLMs. This difficulty appears to stem primarily from the nuanced interpretation required when relating a piece of text to the often abstract goals of an SGD.
\color{black}

Below, we provide representative examples of texts corresponding to each of the three polarity categories, drawn from the OSDG community dataset\footnote{\url{https://zenodo.org/records/5550238}}. This dataset will be further discussed in Section~\ref{sec:Benchmark}.

\begin{itemize}
    \item Positive polarity with respect to SDG‑1 (``End poverty in all its forms everywhere''):

    \begin{itemize}
      \item \textit{``The Brazilian Instituto de Pesquisa Economica Aplicada (Institute of Applied Economic Research, IPEA) has noted that every Real (BRL) invested in the programme increases GDP by BRL 1.44. The 16 million children and adolescents whose school attendance is monitored by the programme show lower rates of truancy and are performing at a level equal to the average student in the public school system, despite their impoverished economic condition. This will lead to a future for these children far different from the situation of exclusion suffered by their parents and grandparents''.}     \end{itemize}

As it can be noted, the text above 
reflects positive progress towards SDG-1 by highlighting how strategic investments in education and economic development can create a virtuous cycle of growth, improved social outcomes, and a break from the longstanding cycles of poverty.

    \item Neutral polarity with respect to SDG‑14 (``Conserve and sustainably use the oceans, seas and marine resources for sustainable development''):

    \begin{itemize}
      \item \textit{``Seafood imports are primarily under the jurisdiction of the Food and Drug Administration. These figures represent total US tariff revenues for imports of edible fish and shellfish products. Since most fishery imports are duty-free, the majority of these amounts are accounted for by imports of a handful of processed products such as canned tuna, sardines and oysters, smoked salmon, and frozen crabmeat. The figures for each year are therefore inflated by approximately 33\%''.} 
    \end{itemize}

This text, on the other hand, simply outlines how seafood import tariff figures are calculated, without providing any insight into the actual management, sustainability, or conservation of marine resources. It focuses on accounting practices and trade data rather than environmental outcomes, so it neither demonstrates progress nor indicates regression towards SDG-14.
    
    \item Negative polarity with respect to SDG‑15 (``Protect, restore and promote sustainable use of terrestrial ecosystems, sustainably manage forests, combat desertification, halt and reverse land degradation, and halt biodiversity loss''):

    \begin{itemize}
      \item \textit{``In MEA and other initiatives the focus has often been on desertification in dry regions and areas in the tropics, far from the Nordic countries. Degraded land is however indeed also present in the Nordic region and concern have risen during the last decades as increased pressures on nature values and biodiversity have been documented in most ecosystems, although at different levels and intensities''.} 
    \end{itemize}

In contrast, this text shows a regression in relation to SDG-15 because it reveals that existing initiatives have focused on desertification in traditionally vulnerable regions, neglecting land degradation issues in the Nordic countries. The rising concerns over increasing pressures on biodiversity and nature in these regions indicate that terrestrial ecosystems are deteriorating, that is a trend contrary to the cited SDG.
    
\end{itemize}

We would like to emphasise that the characteristics of this task do not necessarily coincide with those of the standard sentiment task. For instance, it is possible that a text exhibits characteristics typical of ``positive'' sentiment, but indicates a regression with respect to a specific SDG. This is the case in a text such as:

\textit{``With enthusiasm and hope, we celebrate every daily achievement, while acknowledging that our progress towards Sustainable Development Goal 13 (Climate Action) has fallen behind expectations.''}

In other cases, the text exhibits characteristics typical of ``negative'' sentiment, but actually indicates progress with respect to a specific SDG, as in a sentence such as:

\textit{``Despite significant improvements in access to quality water resources (Sustainable Development Goal 6: Clean Water and Sanitation), the persistent sense of injustice and inequality casts a shadow of deep bitterness.''}

\section{A LLM-based solution for training data generation}
\label{sec:Benchmark}


\color{black}
In this paper, we pursue two main objectives: 1) to investigate the task of SDG polarity detection, and 2) to evaluate whether a synthetic training set, generated using a simple agentic architecture composed of multiple LLM-based annotators, can serve as a high-quality resource for fine-tuning models. The second objective is motivated by the increasing use of synthetic data generated by LLMs in recent years. This approach has been particularly valuable in specialized domains and tasks that suffer from a lack of annotated data. To address both objectives, we developed and publicly released a new benchmark, \textit{SDG-POD}. The training set of \textit{SDG-POD} was generated using the proposed agentic architecture, while the test set was independently annotated by human experts in accordance with established annotation guidelines.
\color{black}

The remainder of this section describes the methodology followed in the creation of the SDG-POD benchmark. We begin by presenting the source data used for building the dataset, then introduce the agentic architecture employed to generate the synthetic training set, and conclude with an explanation of the procedure adopted to create the human-annotated test set.

\subsection{Source data}

SDG-POD includes documents derived from a subset of the dataset released in October 2023 as part of the OSDG community dataset\footnote{\url{https://zenodo.org/records/5550238}}, developed by the OSDG initiative. This dataset is the product of a collaborative effort by more than one thousand volunteers from over 110 countries and contains more than 40,000 text excerpts classified by human volunteers with respect to 16 SDGs.
Each excerpt is 3 to 6 sentences long, averaging about 90 words, and is sourced from publicly available documents such as reports, policy documents, and publication abstracts. A significant portion of these documents, over 3,000, originates from UN-related sources such as SDG-Pathfinder\footnote{\url{https://sdg.iisd.org/news/oecd-tool-applies-sdg-lens-to-international-organizations-policy-content/}} and SDG Library\footnote{\url{https://www.sdglibrary.ca/}}.

To construct SDG-POD, a total of 6,400 texts were randomly sampled from the OSDG dataset, with 400 texts selected for each SDG. This collection was then divided into two subsets: a training set containing 5,824 texts (364 per SDG), which was automatically annotated using LLMs, and a test set comprising 576 texts (36 per SDG), which was evaluated by human annotators.

\subsection{Synthetic training set generation with multiple LLM annotators.}

The training dataset was automatically labelled using a majority voting system based on the classification provided by five LLMs of different features and sizes. 





The five models chosen were as follows:

\begin{enumerate}
    \item  \textit{Meta-Llama-3.1-8B-Instruct:}
        Meta's LLaMa-3.1 is a decoder-only Transformer that comprises a series of generative text models, which have been both pre-trained and fine-tuned. These models range in size from 8 billion to 405 billion parameters. They operate within an auto-regressive framework and leverage an optimized transformer architecture, enabling efficient language processing. The fine-tuning procedure employs both Supervised Fine-Tuning (SFT) and Reinforcement Learning with Human Feedback (RLHF), ensuring that the models align well with human preferences for helpfulness and safety. Additionally, LLaMa-3.1 supports a context length of 128K tokens and demonstrates robust reasoning capabilities. To further enhance its performance in chat and dialogue applications, Meta introduced an innovative instruction fine-tuning approach that combines supervised fine-tuning with techniques such as rejection sampling, proximal policy optimization (PPO), and direct preference optimization (DPO)\footnote{\url{https://ai.meta.com/blog/meta-llama-3-1/)}}.\\

    \item  \textit{Mixtral-8x7B-Instruct-v0.1:}
        Building on the success of its inaugural model, named Mistral, Mistral AI has unveiled its second language model, named Mixtral~\cite{jiang2024mixtral}. This high-quality sparse mixture of experts (SMoE) features open weights and is available under the Apache 2.0 license. Mixtral-8x7B retains key characteristics of its predecessor, including Sliding Window Attention with a training context of 8,000 tokens and Grouped Query Attention, which enhances inference speed. Additionally, it continues to use the Byte-fallback BPE tokenizer for precise character recognition. The Mixtral-8x7B-Instruct-v0.1 variant has been fine-tuned for chat-based inference applications, optimizing its performance for instruction-driven interactions.\\ 

    \item  \textit{Phi-3-mini-4k-instruct:}

        The Phi-3 series~\cite{abdin2024phi3}, launched by Microsoft in April 2024, is a set of compact, decoder-only language models that combine affordability with high performance in language processing, reasoning, coding, and mathematics, often outperforming larger models. Notably, the Phi-3-mini offers both 4K and an unprecedented 128K token context length without significant quality loss, making it ideal for analyzing extensive texts. Developed under Microsoft’s Responsible AI Standard, these models underwent rigorous safety assessments, red-teaming, and enhancements such as reinforcement learning from human feedback. Their small size also makes them well-suited for resource-constrained environments, including on-device and offline applications. Additionally, the Phi-3-mini-4k-instruct variant, with its 4-bit OmniQuant quantization, brings powerful AI chatbot capabilities to devices like iPhones, iPads (with at least 6GB of RAM), and Macs.\\
        
    \item  \textit{Gemma-1.1-7b-it:}

        Gemma~\cite{gemmateam2024gemmaopenmodelsbased} is a series of lightweight, open models developed by Google DeepMind and other Google teams, leveraging the same foundational technology as the Gemini models. They build on recent advances in sequence modeling, transformers, and large-scale distributed training, while also drawing from Google’s legacy of influential open models like Word2Vec, BERT, and T5. These models, trained on 8,192-token contexts, incorporate enhancements such as Multi-Query Attention, rotary positional embeddings, GeGLU activations, and RMSNorm for improved stability.\\

    \item  \textit{Qwen2.5-7B-instruct:}

        The Qwen models~\cite{qwen2025qwen25technicalreport}, developed by Alibaba Cloud, represent a family of LLMs with significant advancements in both pre-training and post-training. Qwen 2.5 introduces substantial improvements over the previous version, starting with an expanded pre-training dataset, scaling from 7 trillion tokens to 18 trillion tokens, enhancing the model’s common sense, expert knowledge, and reasoning abilities. Additionally, the post-training process includes a rigorous supervised fine-tuning with over 1 million samples, along with multistage reinforcement learning, which refines human preference alignment and significantly boosts performance in long text generation, structural data analysis, and instruction execution.

\end{enumerate}

The following prompt was used across all LLMs to perform the classification task using a zero-shot approach:

\begin{verbatim}
f"""Given the following input text, between triple quotes, with its 
associated classification label with respect to one of the Sustainable 
Development Goals (SDGs), further classify the text with respect to 
the three labels defined below:

  - "positive": the text implies or explicitly states that what is 
  affirmed or described leads to a significant advancement in favor 
  of the goal indicated by the SDG under which it was previously 
  classified.

  - "negative": the text implies or explicitly states that what is 
  affirmed or described leads to a significant advancement against 
  the goal indicated by the SDG under which it was previously 
  classified.

  - "neutral": the text does not imply or explicitly state that 
  what is affirmed or described leads to a significant advancement 
  either in favor or against the goal indicated by the SDG under 
  which it was previously classified.

Return the result in JSON format with 4 keys:

  - "label": The assigned label for the input text [this key/value 
  pair is mandatory].

  - "explanation\_1": A description of the reasoning behind the 
  classification decision [this key/value pair is mandatory].

  - "explanation\_2": Additional description of the reasoning 
  behind the classification decision [this key/value pair is 
  optional].

  - "explanation\_3": Another additional description of the 
  reasoning behind the classification decision [this key/value 
  pair is optional].

input = "INPUT TEXT: {input_text}"
SDG\_CLASSIFICATION: "SDG-{sdg}"
"""
\end{verbatim}



To assign a single polarity classification to each item based on the outputs of five LLMs across three polarity categories (positive, negative, and neutral), we adopted a set of carefully designed heuristics.

The distribution of label agreements among the five models was as follows:

\begin{itemize}
\item 774 texts received identical labels from all five models. We refer to this level of agreement as ``Platinum''.
\item 2,338 texts were labelled identically by four out of the five models. This case is denoted as ``Gold''. 
\item 2,429 texts showed agreement among three models, which we label as ``Silver''.
\item 283 texts were classified with only two matching labels. We refer to this lower agreement level as ``Bronze''. \end{itemize}

In the ``Platinum'', ``Gold'', and ``Silver'' cases, the assignment of the final label is straightforward, as a clear majority label emerges. However, the ``Bronze'' cases require a more nuanced approach, as no label reaches a simple majority. To address these cases, we implemented the following  rules:

\begin{itemize}
\item When two models predicted a positive label, two a negative label, and one a neutral label, the final label was set to neutral.
\item When two models predicted a positive label, two a neutral label, and one a negative label, the final label was set to positive.
\item When two models predicted a negative label, two a neutral label, and one a positive label, the final label was set to negative.
\end{itemize}







This procedure led to the creation of a training dataset with the following characteristics:

\begin{itemize}
    \item 2,218 texts labeled as positive (of which 426 ``Platinum'', 888 ``Gold'', 784 ``Silver'', and 120 ``Bronze'')
    \item 2,757 texts labeled as neutral (of which 287 ``Platinum'', 1,212 ``Gold'', 1,221 ``Silver'', and 37 ``Bronze'')
    \item 849 texts labeled as negative (of which 61 ``Platinum'', 238 ``Gold'', 424 ``Silver'', and 126 ``Bronze'')
\end{itemize}


\subsection{Test set construction with human annotators.}

The test dataset of SDG-POD was manually annotated by a team of six human evaluators, divided into two groups of three members. Each group was assigned 288 texts for annotation. To enable the computation of inter-annotator agreement, 48 texts were annotated in common by all members within each group. The remaining 240 texts were evenly divided, with each member independently annotating 80 unique texts not shared with the others in their group. 
A majority voting system was used to determine the final labels for the 48 texts annotated in common. These shared texts were also employed to calculate the inter-annotator agreement and the Cohen’s Kappa coefficient within each group. The purpose of this procedure was to assess the consistency among evaluators. 


The agreement values for each group are presented in Table~\ref{tab:agreement}, while the corresponding Cohen's Kappa coefficients are reported in Table~\ref{tab:Cohen}.



    

    

\begin{table*}[!h]
  \caption{Agreement values measured within each group.}
  \label{tab:agreement}
  \begin{tabular}{l|c|c}
    \toprule
    \textbf{Group} & \textbf{Evaluators} & \textbf{Agreement} \\
    \midrule            
        Group 1 & Evaluator 1 vs. Evaluator 2 &  0.78 \\
        Group 1 & Evaluator 2 vs. Evaluator 3 &  0.58 \\
        Group 1 & Evaluator 3 vs. Evaluator 1 &  0.68 \\
     \midrule            
        Group 2 & Evaluator 4 vs. Evaluator 5 &  0.85 \\
        Group 2 & Evaluator 5 vs. Evaluator 6 &  0.96 \\
        Group 2 & Evaluator 6 vs. Evaluator 4 &  0.88 \\
  \bottomrule
\end{tabular}
\end{table*}



    

    

\begin{table*}[!h]
  \caption{Cohen’s Kappa coefficient values measured within each group.}
  \label{tab:Cohen}
  \begin{tabular}{l|c|c}
    \toprule
    \textbf{Group} & \textbf{Evaluators} & \textbf{Cohen’s Kappa coefficient} \\
    \midrule            
        Group 1 & Evaluator 1 vs. Evaluator 2 &  0.60 \\
        Group 1 & Evaluator 2 vs. Evaluator 3 &  0.38 \\
        Group 1 & Evaluator 3 vs. Evaluator 1 &  0.53 \\
     \midrule            
        Group 2 & Evaluator 4 vs. Evaluator 5 &  0.55 \\
        Group 2 & Evaluator 5 vs. Evaluator 6 &  0.69 \\
        Group 2 & Evaluator 6 vs. Evaluator 4 &  0.57 \\
  \bottomrule
\end{tabular}
\end{table*}

In this case as well, it was necessary to apply a majority voting mechanism to choose the labels for the shared portion among the evaluators. Since the evaluators worked in groups of 3, the possible cases of label combinations in the shared part of the test dataset and the corresponding distributions of texts were as follows:

\begin{itemize}
    \item 59 texts were classified with 3 labels of the same type by the 3 evaluators (this classification was called ``Gold'')
    \item 35 texts were classified with 2 labels of the same type by the 3 evaluators (this classification was called ``Silver'')
    \item 2 texts were classified with 3 different labels by the 3 evaluators (this classification was called ``Bronze'')
\end{itemize}

As with the majority voting mechanism applied to the LLMs, while in the ``Gold'' and ``Silver'' cases the assignment of the final label is straightforward, since one label appears more frequently than the others, in the ``Bronze'' case it was decided that the most appropriate label would be the neutral one. 

\color{black}
The final version of the test set includes 220 texts labeled as positive, 219 as neutral, and 137 as negative.
\color{black}






\section{Experimental Methodology}\label{sec:Methodology}



The analysis presented in this paper has two main objectives. First, to demonstrate that modern LLMs can effectively perform the task of SDG polarity detection, especially if fine-tuned on relevant data. 
Second, to show that a training set composed of synthetically generated data from LLMs can be used to fine-tune models and enhance their performance.

To achieve these goals, we designed an experiment to evaluate the performance of various LLMs in polarity detection, both in ZSL setting and after fine-tuning on synthetic data from the SDG-POD benchmark. The underlying hypothesis is that if the fine-tuned models outperform their non-fine-tuned counterparts, this would indicate that the synthetic training data provides a valuable contribution to the classification task.
\color{black}
The fine-tuning procedure was standardised for all models by employing five training epochs and applying the same prompt as in the zero-shot setting. \color{black}

We selected a set of six language models that are different from those used to generate the synthetic training set. These models are described in detail below.


\begin{enumerate}
    \item  \textit{QwQ-32B-Preview-unsloth-bnb-4bit:}

    Alibaba Cloud's Qwen models form a robust suite of LLMs, with Qwen 2.5 representing a significant leap forward\footnote{\url{https://qwenlm.github.io/blog/qwq-32b-preview/}}. The upgrade expands the pre-training dataset from 7 trillion to 18 trillion tokens, enhancing common sense, expert knowledge, and reasoning capabilities. Post-training improvements include extensive supervised fine-tuning with over one million samples and multi-stage reinforcement learning, which boost long text generation, structural data analysis, and instruction following. The model excels in mathematics, programming, and scientific benchmarks like GPQA and MATH-500. Additionally, the QwQ-32B-Preview is a causal language model built on advanced transformer architecture. It incorporates features such as Rotary Positional Embedding, SwiGLU, RMSNorm, and Attention QKV bias, with 64 layers and 40 attention heads to support deep reasoning. Its extended context length of 32,768 tokens allows it to handle large inputs and complex, multi-step problems.\\

\item  \textit{Phi-4:}

    Phi-4\footnote{\url{https://huggingface.co/microsoft/phi-4}} is a 14B parameter open model from Microsoft, trained on a blend of synthetic datasets, curated public domain content, and academic Q\&A sources. Designed for high-quality reasoning capabilities, the model underwent extensive fine-tuning, including supervised training and direct preference optimization to improve instruction adherence and safety. The training data, an extension of the data used for Phi-3, includes a variety of sources, such as educational materials, code, synthetic "textbook-like" content, and high-quality chat data reflecting human preferences. Multilingual data comprises approximately 8\% of the total dataset, focusing on content that enhances reasoning abilities and correct knowledge alignment.\\

\item  \textit{Mistral-Nemo-Instruct-2407:}

    Mistral-Nemo-Instruct-2407\footnote{\url{https://mistral.ai/news/mistral-nemo}} is a 12B parameter instruct model with a 128k context length, jointly developed by Mistral AI and NVIDIA. It significantly outperforms comparable models in reasoning, world knowledge, and coding, thanks to advanced fine-tuning and alignment. Designed for global, multilingual use, it supports FP8 inference through quantisation awareness and features the new Tekken tokenizer, which compresses text more efficiently than previous tokenizers. Overall, it excels in following precise instructions, managing multi-turn conversations, and generating code.\\

\item  \textit{BERT:}

    BERT~\cite{DBLP:conf/naacl/DevlinCLT19}, introduced by Google in 2018, is an encoder-only Transformer model that set new standards in NLP tasks like language comprehension, question answering, and named entity recognition. It comes in two main variants: BERTbase (12 layers, 12 attention heads, 110 million parameters) and BERTlarge (24 layers, 16 attention heads, 340 million parameters). Trained over four days on a massive corpus from Wikipedia and the Google Books Corpus (https://www.english-corpora.org/googlebooks/), BERT employs a bidirectional masked language modeling technique, masking 15\% of words to predict them based on context, which, combined with transfer learning, enables effective fine-tuning for specific tasks.\\

\item  \textit{ESG-BERT:}

    ESG-BERT~\cite{DBLP:journals/corr/abs-2203-16788} is a language model specifically engineered for text mining in sustainable investing. Leveraging the BERT architecture, it has been fine-tuned to accurately identify and classify content related to Environmental, Social, and Governance (ESG) issues. In classification tasks, it outperforms the standard BERT-base model by achieving a 90\% F1-score compared to 79\%. With 110 million parameters, ESG-BERT was rigorously trained using domain-specific data, reaching 100\% accuracy in Next Sentence Prediction and 98\% in Masked Language Modeling. Additionally, it supports classification across 26 ESG-related categories, from Business Ethics to GHG Emissions.\\



\end{enumerate}

All models were evaluated on the manually annotated test set of SDG-POD using standard evaluation metrics for single-label, multi-class classification: precision, recall, and F1 score.





\section{Results}
\label{sec:results}

Table \ref{tab:ZS} reports the results of the experiments conducted on the SDG-POD test set under the zero-shot learning (ZSL) setting. BERT and ESG-BERT are excluded from this evaluation, as they require fine-tuning and are therefore incompatible with a zero-shot setup.

\begin{table*}[!t]
  \caption{Results of the experiments when employing ZSL. Values are in percentages. In bold the best results.}
  \label{tab:ZS}
  \begin{tabular}{l|rrrr}
    \toprule
    \textbf{MODEL NAME} & \textbf{Pre} & \textbf{Rec} & \textbf{Acc} & \textbf{F1} \\
    \midrule            
        PHI4-4B &  62.0 &  \textbf{62.8} & 61.3 &  59.8 \\  
        MISTRAL-NEMO-12B & 59.5 & 61.3 & 59.5 & 57.7 \\ 
        QWQ-32B & 59.0 & 59.9 & 59.2 & 57.8 \\    
  \bottomrule
\end{tabular}
\end{table*}

\begin{table*}[!t]
  \caption{Results of the experiments when employing FT. Values are in percentages. In bold the best results. }
  \label{tab:FT}
  \centering
    \begin{tabular}{l|rrrr}
    \toprule
    \textbf{MODEL NAME} & \textbf{Pre} & \textbf{Rec} & \textbf{Acc} & \textbf{F1} \\
    \midrule      
        BERT & 56.8 & 51.8 & 54.5 & 52.6 \\
        ESG-BERT & 58.9 & 56.5 & 57.8 & 57.2 \\
        PHI4-4B &  65.4 & 59.9 & 61.8 & 61.3 \\  
        MISTRAL-NEMO-12B & 63.8 & 58.8 & 60.6 & 60.2 \\ 
        QWQ-32B & \textbf{66.5} & \textbf{60.1} & \textbf{62.0} & \textbf{61.6} \\ 
  \bottomrule
\end{tabular}
\end{table*}

\begin{figure}[b!]
    \centering
    \includegraphics[width=13cm]{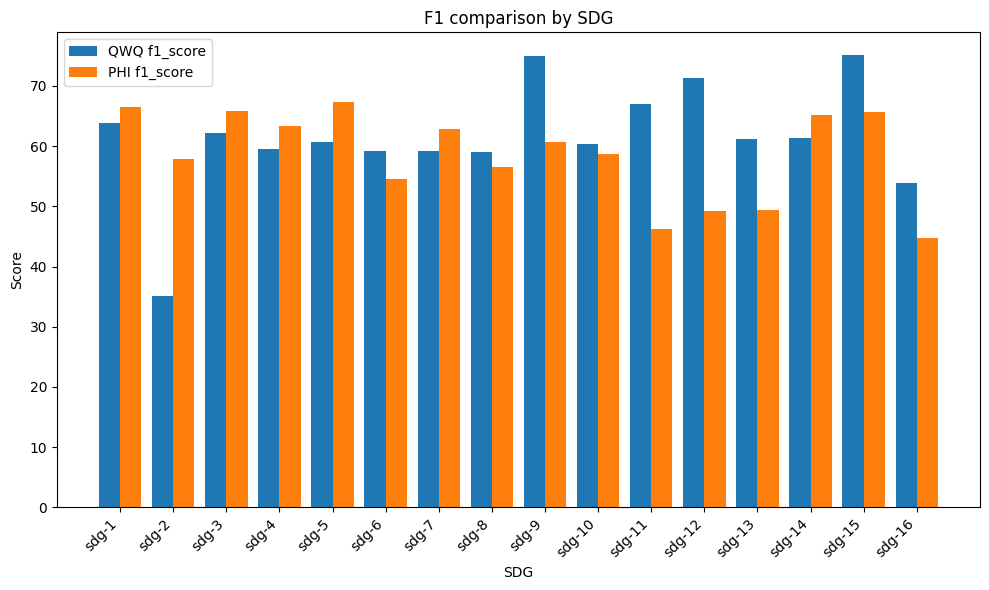}
    \caption{F1 comparison by SDG for QWQ-32B and Phi4-8B LLMs.}
    \label{fig:F1_comparison_by_SDG}
\end{figure}

In the case of zero-shot experiments, it was observed that the best open source performing model was the Phi4 model, which has the smallest number of parameters (only 4B), with a measured F1 score of 59.8\%. 
The Phi4 good result was achieved thanks to its optimized architecture and efficient training methodologies, which provide a competitive advantage over other models, even those with a larger number of parameters, in situations with limited available information.

Table \ref{tab:FT} instead shows the results of the classification experiments conducted on the models after they were fine-tuned. Unlike before, in this case the best open source performing model was the one with the largest number of parameters (QWQ-32B), \color{black} although the Phi-4 model was fine-tuned for 10 epochs (whereas QWQ-32B was trained for only 5) in order to improve its classification metrics, \color{black} with a measured F1-score of 61.6\%, an increase of 3.8 percentage points compared to the performance of the non-fine-tuned version of the same model and a gap of 1.8 percentage points compared to the performance of the best non-fine-tuned model. 

Although the overall F1-score gains between zero-shot and fine-tuned settings may seem modest at first glance,
a closer examination reveals the substantial benefits of fine-tuning, particularly in reducing critical errors and improving reliability across specific SDG classes.

\begin{table}[!b]
  \centering
  \caption{PHI-4 Zero-Shot.}
  \label{tab:phi-4_zs_conf_matr}
  \begin{tabular}{l*{9}{c}}
    \toprule
    \textbf{Actual}

      & \multicolumn{3}{c}{\textbf{Positive (220)}}
      & \multicolumn{3}{c}{\textbf{Neutral (219)}}
      & \multicolumn{3}{c}{\textbf{Negative (137)}} \\      
    \cmidrule(lr){2-4}\cmidrule(lr){5-7}\cmidrule(lr){8-10}
    \textbf{Predicted}
    
      & \textbf{Pos} & Neu & Neg
      & Pos & \textbf{Neu} & Neg
      & Pos & Neu & \textbf{Neg} \\
    \midrule

    \textbf{SDG 1}  & \textbf{12} & 2 & 2 & 5 & \textbf{8}  & 2 & 0 & 1 & \textbf{4}  \\
    \textbf{SDG 2}  & \textbf{10} & 4 & 0 & 7 & \textbf{3}  & 1 & 1 & 2 & \textbf{8}  \\
    \textbf{SDG 3}  & \textbf{14} & 0 & 2 & 6 & \textbf{3}  & 2 & 0 & 1 & \textbf{8}  \\
    \textbf{SDG 4}  & \textbf{17} & 1 & 0 & 8 & \textbf{3}  & 0 & 3 & 0 & \textbf{4}  \\
    \textbf{SDG 5}  & \textbf{13} & 0 & 0 & 6 & \textbf{3}  & 3 & 0 & 1 & \textbf{10} \\
    \textbf{SDG 6}  & \textbf{9}  & 2 & 3 & 5 & \textbf{1}  & 3 & 1 & 1 & \textbf{11} \\
    \textbf{SDG 7}  & \textbf{12} & 1 & 1 & 7 & \textbf{8}  & 1 & 1 & 2 & \textbf{3}  \\
    \textbf{SDG 8}  & \textbf{10} & 1 & 1 & 3 & \textbf{1}  & 4 & 2 & 4 & \textbf{10} \\
    \textbf{SDG 9}  & \textbf{18} & 0 & 0 & 11 & \textbf{3} & 0 & 1 & 0 & \textbf{3}  \\
    \textbf{SDG 10} & \textbf{6}  & 0 & 2 & 5 & \textbf{6}  & 6 & 0 & 1 & \textbf{10} \\
    \textbf{SDG 11} & \textbf{11} & 1 & 0 & 10 & \textbf{1} & 2 & 3 & 1 & \textbf{7}  \\
    \textbf{SDG 12} & \textbf{11} & 1 & 1 & 6 & \textbf{3}  & 5 & 1 & 3 & \textbf{5}  \\
    \textbf{SDG 13} & \textbf{12} & 2 & 1 & 9 & \textbf{4}  & 3 & 1 & 1 & \textbf{3}  \\
    \textbf{SDG 14} & \textbf{10} & 1 & 0 & 8 & \textbf{8}  & 2 & 0 & 1 & \textbf{6}  \\
    \textbf{SDG 15} & \textbf{14} & 0 & 1 & 8 & \textbf{5}  & 2 & 0 & 0 & \textbf{6}  \\
    \textbf{SDG 16} & \textbf{6}  & 4 & 1 & 6 & \textbf{7}  & 6 & 2 & 1 & \textbf{3}  \\

    \midrule

    \textbf{SUM} &  \textbf{185} & 20 &  15 &  110 & \textbf{67}  &  42 &  16 & 20 &  \textbf{101} \\  
    
    \bottomrule
  \end{tabular}
\end{table}

\begin{table}[!t]
  \centering
  \caption{QWQ Fine-Tuned.}
  \label{tab:qwq_ft_conf_matr}
  \begin{tabular}{l*{9}{c}}
    \toprule
    \textbf{Actual}

      & \multicolumn{3}{c}{\textbf{Positive (220)}}
      & \multicolumn{3}{c}{\textbf{Neutral (219)}}
      & \multicolumn{3}{c}{\textbf{Negative (137)}} \\       
    \cmidrule(lr){2-4}\cmidrule(lr){5-7}\cmidrule(lr){8-10}
    \textbf{Predicted}
      & \textbf{Pos} & Neu & Neg
      & Pos & \textbf{Neu} & Neg
      & Pos & Neu & \textbf{Neg} \\
    \midrule

    \textbf{SDG 1}  & \textbf{10} & 5 & 1 & 3 & \textbf{11} & 1 & 0 & 3 & \textbf{2}  \\
    \textbf{SDG 2}  & \textbf{4}  & 10 & 0 & 6 & \textbf{5}  & 0 & 0 & 8 & \textbf{3}  \\
    \textbf{SDG 3}  & \textbf{12} & 3 & 1 & 2 & \textbf{9}  & 0 & 0 & 7 & \textbf{2}  \\
    \textbf{SDG 4}  & \textbf{17} & 1 & 0 & 4 & \textbf{7}  & 0 & 1 & 6 & \textbf{0}  \\
    \textbf{SDG 5}  & \textbf{11} & 2 & 0 & 5 & \textbf{6}  & 1 & 0 & 6 & \textbf{5}  \\
    \textbf{SDG 6}  & \textbf{5}  & 8 & 1 & 2 & \textbf{7}  & 0 & 0 & 4 & \textbf{9}  \\
    \textbf{SDG 7}  & \textbf{11} & 3 & 0 & 4 & \textbf{10} & 2 & 1 & 4 & \textbf{1}  \\
    \textbf{SDG 8}  & \textbf{8}  & 4 & 0 & 2 & \textbf{4}  & 2 & 0 & 8 & \textbf{8}  \\
    \textbf{SDG 9}  & \textbf{15} & 3 & 0 & 5 & \textbf{9}  & 0 & 0 & 1 & \textbf{3}  \\
    \textbf{SDG 10} & \textbf{3}  & 4 & 1 & 3 & \textbf{13} & 1 & 0 & 5 & \textbf{6}  \\
    \textbf{SDG 11} & \textbf{9}  & 3 & 0 & 4 & \textbf{9}  & 0 & 2 & 3 & \textbf{6}  \\
    \textbf{SDG 12} & \textbf{11} & 2 & 0 & 3 & \textbf{11} & 0 & 1 & 4 & \textbf{4}  \\
    \textbf{SDG 13} & \textbf{10} & 4 & 1 & 4 & \textbf{10} & 2 & 0 & 3 & \textbf{2}  \\
    \textbf{SDG 14} & \textbf{6}  & 5 & 0 & 7 & \textbf{10} & 1 & 0 & 1 & \textbf{6}  \\
    \textbf{SDG 15} & \textbf{12} & 3 & 0 & 4 & \textbf{11} & 0 & 0 & 2 & \textbf{4}  \\
    \textbf{SDG 16} & \textbf{3}  & 8 & 0 & 3 & \textbf{14} & 2 & 2 & 1 & \textbf{3}  \\

    \midrule

    \textbf{SUM} &  \textbf{147} & 68 &  5 &  61 & \textbf{146}  &  12 &  7 & 66 &  \textbf{64} \\   
    \bottomrule
  \end{tabular}
\end{table}

\begin{table}[ht]
  \caption{Macro and Micro AVG F1 comparison.}

  \centering
  \begin{minipage}{0.48\textwidth}
    \centering
    \textbf{MACRO AVG F1}\\[1ex]
    \footnotesize
    \begin{tabular}{@{}lc@{}}
    
      \toprule
      \multicolumn{2}{c}{\emph{cost\_pn=cost\_np=1}} \\
      
      \midrule
      LLM               & Weighted F1 \\
      \midrule
      Phi4 (ZS)    & 59.8  \\
      Mistral (ZS) & 57.7  \\
      QWQ (ZS)     & 57.8  \\
      Phi4 (FT)    & 61.3  \\
      Mistral (FT) & 60.2  \\
      QWQ (FT)     & 61.6  \\
      \midrule
      \multicolumn{2}{c}{\emph{cost\_pn=cost\_np=2}} \\
      
      \midrule
      LLM               & Weighted F1 \\
      \midrule
      Phi4 (ZS)    & 56.3  \\
      Mistral (ZS) & 53.7  \\
      QWQ (ZS)     & 54.7  \\
      Phi4 (FT)    & 59.6  \\
      Mistral (FT) & 58.7  \\
      QWQ (FT)     & 60.0  \\
      \midrule
      \multicolumn{2}{c}{\emph{cost\_pn=cost\_np=10}} \\
      
      \midrule
      LLM               & Weighted F1 \\
      \midrule
      Phi4 (ZS)    & 40.7  \\
      Mistral (ZS) & 37.1  \\
      QWQ (ZS)     & 40.2  \\
      Phi4 (FT)    & 49.9  \\
      Mistral (FT) & 50.2  \\
      QWQ (FT)     & 50.6  \\
      \bottomrule
    \end{tabular}
  \end{minipage}
  \hfill
  \begin{minipage}{0.48\textwidth}
    \centering
    \textbf{MICRO AVG F1}\\[1ex]
    \footnotesize
    \begin{tabular}{@{}lc@{}}
    
      \toprule
      \multicolumn{2}{c}{\emph{cost\_pn=cost\_np=1}} \\      
      \midrule
      LLM               & Weighted F1 \\
      \midrule
      Phi4 (ZS)    & 61.3  \\
      Mistral (ZS) & 59.5  \\
      QWQ (ZS)     & 59.2  \\
      Phi4 (FT)    & 61.8  \\
      Mistral (FT) & 60.6  \\
      QWQ (FT)     & 62.0  \\    
      \midrule
      \multicolumn{2}{c}{\emph{cost\_pn=cost\_np=2}} \\      
      \midrule
      LLM               & Weighted F1 \\
      \midrule
      Phi4 (ZS)    & 58.2  \\
      Mistral (ZS) & 55.8  \\
      QWQ (ZS)     & 56.5  \\
      Phi4 (FT)    & 60.4  \\
      Mistral (FT) & 59.5  \\
      QWQ (FT)     & 60.7  \\       
      \midrule
      \multicolumn{2}{c}{\emph{cost\_pn=cost\_np=10}} \\
      \midrule
      LLM               & Weighted F1 \\
      \midrule
      Phi4 (ZS)    & 41.3  \\
      Mistral (ZS) & 37.0  \\
      QWQ (ZS)     & 41.2  \\
      Phi4 (FT)    & 51.4  \\
      Mistral (FT) & 51.7  \\
      QWQ (FT)     & 52.2  \\     
      \bottomrule
    \end{tabular}
  \end{minipage}
  \label{tab:weighted_f1}
\end{table}

Figure \ref{fig:F1_comparison_by_SDG} shows a side-by-side comparison of F1-scores, broken down by each SDG, for the best zero-shot model (Phi-4) and the best fine-tuned model (QWQ). When Phi-4 edges out QWQ, the margin is almost negligible, with the lone exception of SDG-2. In contrast, when QWQ outperforms Phi-4, the gaps are substantially larger. This pattern underscores that fine-tuning can deliver pronounced gains in specific cases.

Beyond the quantitative comparison, the results suggest that the intrinsic characteristics of individual SDGs play an important role in determining classification difficulty. Some goals, such as SDG-9 (Industry, Innovation and Infrastructure), SDG-12 (Responsible Consumption and Production), and SDG-15 (Life on Land), are often associated with relatively explicit domain-specific terminology (e.g., infrastructure projects, recycling practices, biodiversity, or ecosystem degradation), enabling the models to identify the direction of the reported impact more reliably. In contrast, goals such as SDG-2 (Zero Hunger), SDG-10 (Reduced Inequalities), or SDG-16 (Peace, Justice and Strong Institutions) frequently involve more nuanced social, political, and economic contexts. In these cases, determining whether a text describes genuine progress, regression, or merely a factual situation often requires broader contextual reasoning and the interpretation of implicit relationships rather than relying on explicit lexical cues. This observation suggests that SDG polarity detection is not equally difficult across all goals, and that future models should place greater emphasis on contextual and causal reasoning to improve performance on the more challenging SDGs. This variability also highlights the value of reporting per-SDG performance rather than only aggregate metrics, as overall F1 scores can mask substantial differences in task complexity across sustainability domains.

Comparing the confusion matrices for each SDG (Tables \ref{tab:phi-4_zs_conf_matr} and \ref{tab:qwq_ft_conf_matr}) reveals that the non-fine-tuned Phi-4 model frequently mistakes negative examples for positive (and vice versa). The fine-tuned QWQ model, however, commits these ``critical'' errors far less often, at worst confusing a positive or negative label with the neutral class. Since confusing positive and negative labels is more severe than assigning them to neutral, this demonstrates that fine-tuning reduces the most serious misclassifications.

\color{black}
In light of these considerations, we compared the performance of the non-fine-tuned model with that of the fine-tuned version using the error-weighted F1 metric, which is commonly employed when certain types of misclassifications are more critical than others~\citep{NIPS2014_5c0314ec}. Specifically, we assigned heavier penalties (twofold and tenfold) to severe errors involving the incorrect prediction of negative labels as positive and vice versa, in order to emphasize which model was more robust to such critical mistakes. Table~\ref{tab:weighted_f1} reports both macro and micro weighted F1 scores. The results indicate that as larger weights are applied to the most consequential errors, the F1-score gap between the non-fine-tuned and fine-tuned models increases markedly, from approximately 1–2 percentage points to about 10–11. This trend underscores the substantial performance improvements yielded by fine-tuning.

The evaluation indicates that fine-tuning on synthetic data generated with the novel technique proposed in this paper, which integrates the outputs of multiple LLMs, leads to consistently improved results. The models obtained not only achieve higher overall performance but also demonstrate greater robustness, particularly by reducing the occurrence of severe misclassifications where polarity is reversed. It also highlights that the task of SDG polarity classification remains highly challenging and far from solved. Given the importance of monitoring SDG concepts, this task constitutes a valuable benchmark for future systems and an important avenue for further exploration.
\color{black}
\color{black}

\color{black}
Moreover, in addition to the analyses described above, we also applied the McNemar statistical test to compare the performance of our fine-tuned models against the baseline. The test reveals that, for the positive class, the improvements achieved by both Phi4 and Mistral are statistically significant, with p-values of 0.04 and 0.05, respectively. For the neutral and negative classes, statistical significance could not be established, which is likely attributable to the relatively smaller number of data points in these categories, limiting the test's sensitivity.

\color{black}

\section{Conclusions}
\label{sec:conclusions}

In this work, we explored the task of polarity detection for SDGs in news text, leveraging LLMs to assess how different architectures and training strategies perform in this classification scenario. Our analysis shows that fine-tuned models substantially outperform their zero-shot counterparts, particularly when we account for the severity of classification errors such as confusing positive and negative sentiments. These improvements are especially evident when using weighted evaluation metrics that penalize critical misclassifications more heavily.

To support this study, we introduced SDG-POD, a benchmark dataset specifically designed for polarity detection in the SDG domain. By combining  manually and automatically annotated data, SDG-POD enables systematic evaluation of LLMs on a task where annotated data is limited and costly to obtain. 


\color{black}
Our experiments indicate that SDG polarity detection remains highly challenging, even for the latest generation of LLMs that achieve strong results on a wide range of NLP tasks. The difficulty stems from the subtle and abstract nature of the text, which is readily interpretable by human experts but considerably less accessible to LLMs. As a result, the SDG-POD represents a demanding benchmark for future research, and we expect that the community will continue to adapt and advance in addressing this task.

As an initial step in this direction, we demonstrated that synthetic data generated with the novel method introduced in this paper, which integrates the outputs of multiple LLMs, yields consistently improved performance. The resulting models not only achieve higher overall accuracy but also exhibit greater robustness, particularly through a reduction in severe misclassifications involving polarity reversal. Notably, QWQ-32B trained on this data surpassed all competing approaches.

Beyond aggregate performance, our per-SDG analysis and confusion matrix comparisons show that fine-tuning also mitigates the most consequential errors. These results underscore the importance of careful model adaptation and targeted evaluation when applying LLMs in socially significant domains such as sustainability.
\color{black}

\color{black}




Overall, this work provides both methodological insights and practical tools for advancing polarity detection in SDG-related news content.

Despite the encouraging results, our approach has several limitations. The synthetic training labels were generated through a majority voting mechanism among multiple LLMs, which reduces the influence of individual model errors but cannot completely eliminate systematic biases shared across models. Consequently, these biases may propagate to the downstream models fine-tuned on the synthetic annotations, potentially affecting their predictions in systematic ways. This issue is particularly relevant in the sustainability domain, where the interpretation of progress towards specific SDGs may vary across geographical regions, institutional contexts, and socio-economic perspectives. Consequently, although synthetic annotation offers a scalable alternative to costly manual labeling, it should not be regarded as a substitute for high-quality human annotation in applications requiring high levels of accountability or policy support. For this reason, the results reported in this paper should be interpreted as demonstrating the effectiveness of synthetic data for model adaptation rather than as evidence that synthetic annotation can fully replace expert-generated labels.

Future work will investigate hybrid annotation strategies that combine synthetic labels with expert validation, active learning, and geographically diverse human annotators to mitigate potential biases while preserving the scalability advantages of LLM-assisted data generation. We also plan to explore the integration of SDG polarity detection into broader SDG mapping frameworks, such as the STRINGS project, where polarity information could provide an additional semantic layer for characterizing the relationship between scientific research and individual SDGs, enabling richer and more nuanced analyses for researchers, policymakers, and other stakeholders. Finally, we intend to expand SDG-POD to include multilingual data and evaluate its applicability in real-world scenarios, including policy monitoring, media analysis, and decision-support systems for sustainability.

\color{black}





\bibliographystyle{elsarticle-num} 
\bibliography{bibliography}



\end{document}